\documentclass[runningheads]{llncs}
\usepackage{graphicx}
\usepackage{enumitem}
\usepackage{cite}
\usepackage{amsmath,amssymb,amsfonts}
\usepackage{algorithmic}
\usepackage{graphicx}
\usepackage{textcomp}
\usepackage{xcolor}

\usepackage{booktabs} 
\usepackage{amssymb}
\usepackage{makecell} 
\usepackage{multirow}
\usepackage{xcolor} 

\usepackage{amsmath}  
\usepackage{bm} 
\begin{document}
\title{ARD-REFSM: Enhancing Reflection Symmetry Detection with Asymmetric Denoising and Rotation Equivariance}
%
%

\author{Dongfu Yin\inst{1} \and
Rourou Su\inst{1,2} \and
Cong Zhao\inst{1,3} \and
Fei Yu\inst{4}}


%

%

\authorrunning{Dongfu Yin et al.} 
\titlerunning{ARD-REFSMNet}

\institute{Guangdong Laboratory of Artificial Intelligence and Digital Economy(SZ), Shenzhen, China \and
College of Computer Science and Software Engineering, Shenzhen University, China \and
Department of Biomedical Engineering, Southern University of Science and Technology, China \and
School of Information Technology, Carleton University, Canada
}


%
\maketitle              
\begin{abstract}
  Reflection symmetry detection remains challenging due to interference from asymmetric regions and arbitrary orientations of symmetric patterns. Asymmetric regions introduce background clutter that disrupts symmetric pattern matching, whereas conventional convolutional neural networks lack rotation equivariance, leading to inconsistent feature representations under rotational transformations. To address these issues, we propose an Asymmetric Region Denoising (ARD) module and a Rotation Equivariant Feature Similarity Matching (REFSM) module. The ARD module suppresses asymmetric interference to refine symmetric patterns, while the REFSM module enhances rotation equivariance through feature similarity matching between original and rotated images. Specifically, our dual-input REFSM framework leverages rotation loss to maximize consistency between the score maps of original and rotated images, thereby enabling precise prediction of rotation-equivariant symmetry axes. Furthermore, we introduce GMSYM, a new benchmark dataset that categorizes images into diverse scenarios and incorporates various interferences to address the limitations of existing reflection symmetry detection benchmarks. Extensive experiments on four standard datasets (DENDI, NYU, LDRS, SDRW) and our proposed GMSYM dataset demonstrate that our method achieves state-of-the-art performance in both accuracy and robustness.

\end{abstract}

\begin{figure}[t]
    \centering
    \includegraphics[width=0.9\columnwidth]{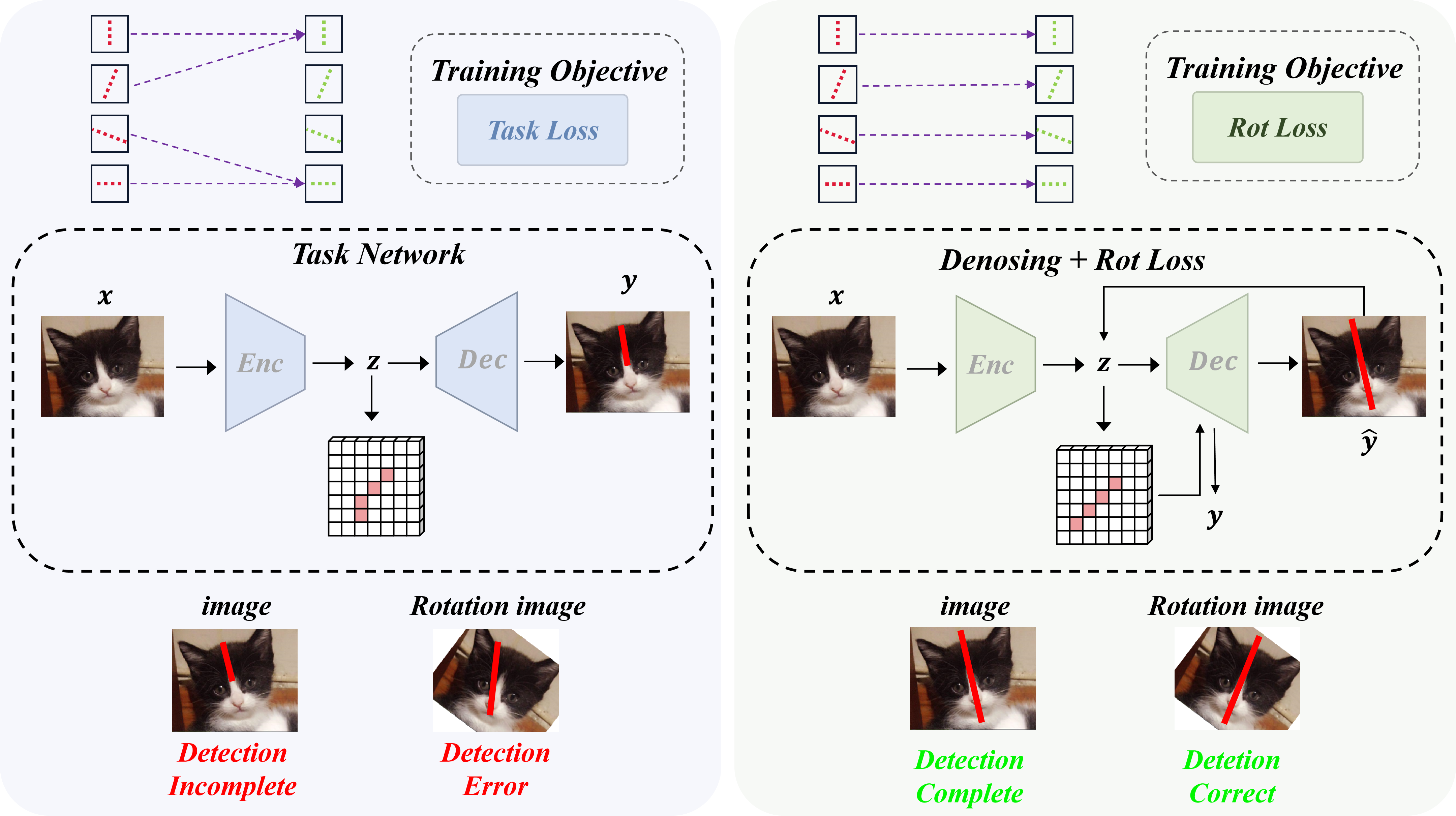} 
    \caption{(a) Most image symmetry detection models exhibit two primary limitations: vulnerability to asymmetric regions interference and frequent failure in detecting symmetries within rotated images, (b) Let $z$ denote features extracted from the original image $x$, and $y$ represent the final prediction, (c) Our method effectively predicts reflection symmetry by incorporating asymmetric region denoising and rotation loss.}
    \label{fig1}
\end{figure}

\section{Introduction}
Symmetry is a fundamental attribute of image content that has garnered significant attention due to its regularity and perceptual salience. Leveraging symmetry, images can be compressed by preserving only half of the information, while the other half can be exploited for image completion, denoising, and restoration. This paper focuses on reflection symmetry detection, which aims to identify the reflection axis that remains invariant under reflection transformations~\cite{Zhou_2021_CVPR,9808406,10378251}.

Existing methods for reflection symmetry detection are mainly categorized into keypoint matching~\cite{DBLP:conf/iccvw/CicconetHE17,ALVARADOGONZALEZ201967} and scoremap prediction~\cite{7913730,DBLP:journals/tip/GnuttiGL21}. Scoremap prediction methods predict symmetry axes by learning pixel-level symmetry score maps from training data~\cite{DBLP:conf/iccvw/FunkLOTSCDL17,DBLP:conf/iccv/FunkL17}, achieving significantly superior performance compared to traditional approaches. State-of-the-art methods for both 2D and 3D symmetry detection~\cite{10377455,Je_2024,DBLP:conf/cvpr/Li0TR25,Zhao_2023_ICCV} are capable of predicting pixel-wise symmetry scores. Despite the mathematical simplicity of symmetry and the substantial body of literature on this topic, detecting symmetric patterns in real-world images remains challenging due to multiple interfering factors. As illustrated in Fig.~\ref{fig1}, interference from asymmetric regions often introduces spurious artifacts near the symmetry axis, resulting in incomplete or inaccurate axis detection~\cite{DBLP:conf/iccv/SeoSC21}. Furthermore, for rotated images, conventional convolutional neural networks fail to maintain consistent feature representations for symmetric patterns due to the lack of rotation equivariance, leading to unreliable symmetry axis predictions~\cite{DBLP:conf/cvpr/SeoKKC22}.

To address these challenges, we propose an asymmetric region denoising technique that leverages the preliminary symmetry score map as spatial weights to guide the network in suppressing asymmetric interference. Additionally, we introduce a rotation equivariant feature similarity matching mechanism to enhance prediction consistency across different orientations. The main contributions of this work are summarized as follows:
\begin{itemize}[noitemsep, topsep=0pt]
    \item We propose ARDNet, a reflection symmetry detection network equipped with an Asymmetric Region Denoising (ARD) module. The ARD module refines symmetric regions by suppressing spurious artifacts in symmetry score maps, yielding more precise symmetric pattern predictions.
    \item We introduce a Rotation Equivariant Feature Similarity Matching (REFSM) module that enhances the consistency between score maps of non-rotated and rotated images, thereby enabling rotation-equivariant symmetry axis prediction.
    \item We construct GMSYM, a new benchmark dataset that categorizes images into diverse scenarios with various interferences, alleviating the limitations of existing reflection symmetry detection benchmarks. Extensive experiments on the DENDI and GMSYM datasets demonstrate that our method achieves state-of-the-art performance in both accuracy and robustness, and the effectiveness of each component is validated through ablation studies.
\end{itemize}

\begin{figure*}[h]
    \centering
    \includegraphics[width=\textwidth]{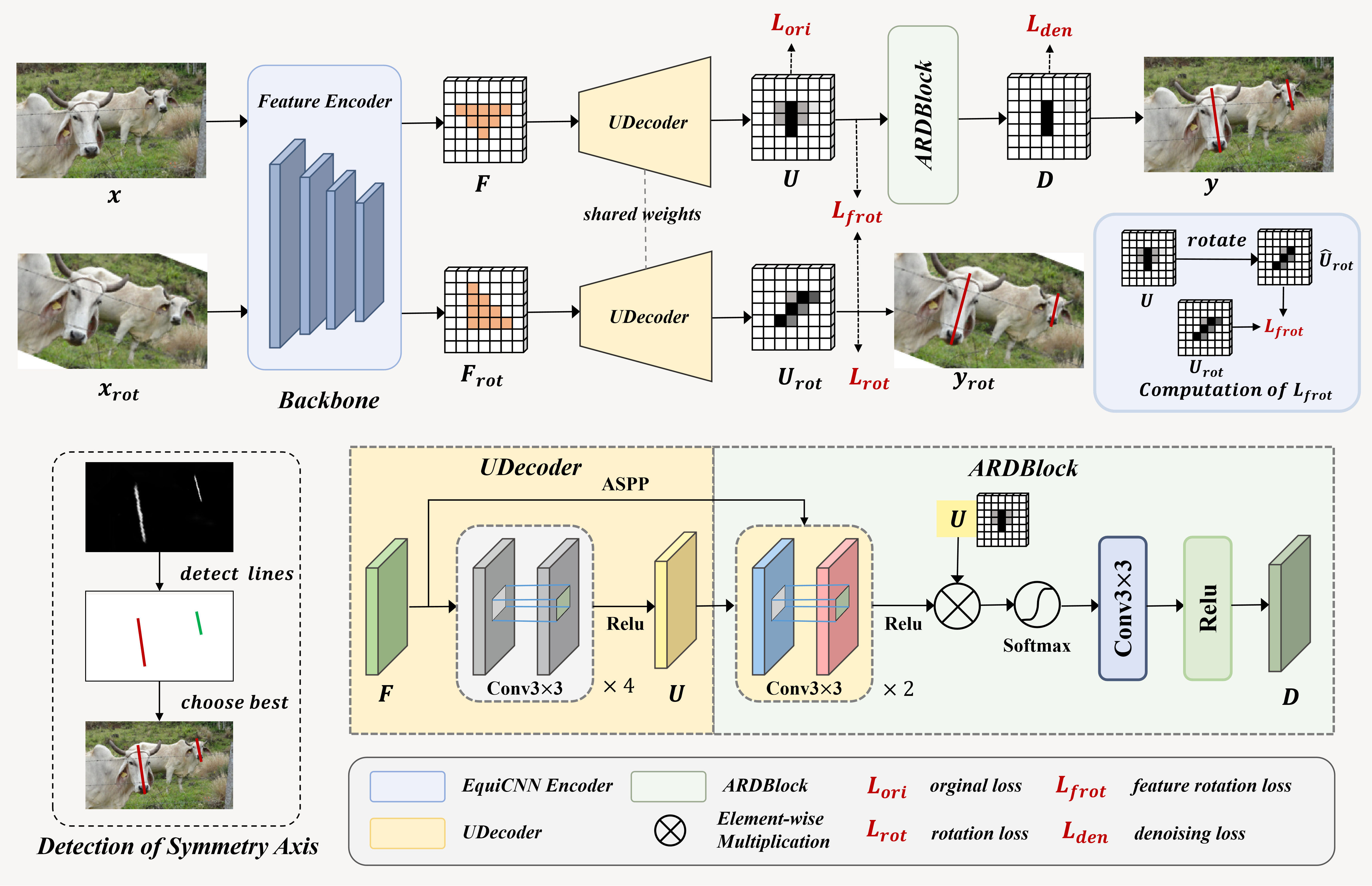}
    \caption{Overview of our proposed ARDNet. Given the input image \(\bm{x}\) without rotation and the image \(  \bm{x_{rot}} \) with random rotation, preliminary symmetric score maps \(\bm{U}\) and  \(  \bm{U_{rot}} \)  are obtained through a group-equivariant encoder and a group-equivariant decoder, respectively. Then, \(\bm{U}\) is passed through a asymmetric region denoising module to obtain the final prediction result \(\bm{D}\). The obtained \(\bm{U}\) undergo the same rotation as \(\bm{x_{rot}}\) to obtain \( \bm{\hat{U}_{rot}} \), maximizing the similarity between \( \bm{\hat{U}_{rot}} \) and \(  \bm{U_{rot}} \) to achieve rotation-invariant prediction.}
    \label{fig2}
    \end{figure*}

    \section{PROPOSED METHOD}
    
    We propose an asymmetric region denoising framework for symmetry detection using equivariant feature similarity matching. Our proposed ARDNet framework is illustrated in Fig. \ref{fig2}.

    \subsection{Asymmetric Region Denoising}
    Existing methods often appear artifacts when detecting regions near the symmetry axis, which interfere with the detection of the background class. Therefore, we propose asymmetric region denoising. The specific method is as follows: given an image, the original image is encoded by the encoder to extract feature maps containing rich information. We adopt the framework of equivariant backbone, which consists of four layers.We represent the input image as $\bm{x}$. Then, the multi-layer features  \( \bm{F_{i}} \) obtained through encoding can be represented as following,
    \begin{equation}
      \bm{F_i} = \bm{Encoder}(Layer_i(\bm{x}))
    \end{equation}
    
    We upsample these four layers of feature maps layer by layer, and finally decode to obtain the final predicted image. This allows us to utilize more image information, resulting in a more accurate symmetric score map \(\bm{U}\) ,  denoted as following,
    \begin{equation}
     \bm{D_i} = \bm{Decoder}(\bm{F_i},\bm{ D_{i+1}})
    \end{equation}
    \begin{equation}
     \bm{U} = \bm{UDecoder}(\bm{F_i},\bm{ D_{i}})
    \end{equation}
     $D_i$ is obtained through the feature map $F_i$ and upsampling decoding, where $i = 1, 2, 3$. $U$ denotes the preliminary symmetry score map generated from encoded feature maps $F_i$ and decoded feature maps $D_i$ for $i=1, 2, 3, 4$. Specifically, $ D_4 = Decoder(F_4) $.
    
    Through the encoder-decoder process, we can obtain a rough symmetry score map by decoding. To further extract multi-scale context, we pass \( \bm{F_{4}} \) through ASPP to obtain feature maps $ \bm{P} $ with richer semantic information. 
    \begin{equation}
      \bm{P} = \bm{ASPP}(\bm{F_4})
    \end{equation}
    After those process, we can obtain an initial symmetric score map \(\bm{U}\), which helps to locate the specific regions where the symmetry axis lies. We regard the symmetric score map as weights, where the foreground class in the score map, namely the symmetric regions, carries more weight than the background class. Using the score map as weights enables the original features to focus more on the symmetric axis regions while ignoring the background, thereby eliminating the interference of artifacts and noise. The implementation of \(\bm{ARD} \) can be expressed as follows, 
    \begin{equation}
      \bm{D} = \bm{ARD}(\bm{P}, \bm{U})
    \end{equation}
    \begin{equation}
      \bm{ARD}(\bm{P}, \bm{U}) = \frac { exp(\bm{P_{h,w,c}} * \bm{U_{h,w,c}}) } {\sum_{c} exp(\bm{P_{h,w,c}} * \bm{U_{h,w,c}})} 
    \end{equation}

    \subsection{Rotation Equivariant Feature Similarity Matching}
    
    Given an original image that has not undergone any rotation transformation, we randomly rotate the original image and generate a rotated version in the range of 0° to 45°.For the score maps of the original image, \( \bm{U} \), and the rotated image, \( \bm{U_{\text{rot}}} \), we aim to increase their similarity to the ground truth.
    
    Therefore, we rotate the score map \(\bm{U}\) of the original image by the same angle as the rotation branch \(\bm{x_{rot}}\), and denote the resulting mapped image as \( \bm{\hat{U}_{rot}} \) . Since the rotated prediction map \( \bm{\hat{U}_{rot}} \) and the rotation branch score map \( \bm{U_{rot}} \) should produce consistent predictions for the same image,  we aim to increase their similarity, which can be expressed as:
    
    \begin{equation}
    \mathop{\bm{argmin}}\limits_{x}(\frac{1}{2} \Vert \bm{U_{rot}}-\bm{U^{gt}_{rot}} \Vert_2^2 + \frac{1}{2} \Vert \bm{\hat{U}_{rot}}-\bm{U_{rot}} \Vert_2^2)
    \end{equation}
    
    where \( x \) represents the parameters of our network.
    
    We randomly rotate the second input from 0° to 45° because we use N=8 rotational equivariant convolutions. Combining these rotations achieves an effect of 45° × 8 = 360°, covering various angles of an image.

    \subsection{Training objective}
    Symmetry detection involves foreground classes representing symmetric axes and background classes representing asymmetric regions, so we employ pixel-level binary classification to detect symmetric axes. As the symmetric axis regions occupy only a small portion and there exists an imbalance with asymmetric regions, we use $\alpha$-variant of the focal loss $\bm{Lf}(\bm{P}_{w,h}, \bm{Y}_{w,h})$ to balance the rate of these two classes.
    
    We utilized focal loss to calculate the original loss \( \bm{L_{ori}} \) for the score map of the original image $\bm{U}$. and compute the rotation loss \( \bm{L_{rot}} \) for the score map of the rotated image $\bm{U^{gt}}$. We employed the score map obtained from the asymmetric region denoising module $\bm{D}$ as our symmetric detection results. To enhance the accuracy of rotation detection, we introduced the feature rotation loss \( \bm{L_{frot}} \). 
    
    \begin{equation}
      \bm{L_{ori}} = \bm{Lf} ( \bm{U}, \bm{U^{gt}})
    \end{equation}
    \begin{equation}
      \bm{L_{rot}} = \bm{Lf} ( \bm{U_{rot}}, \bm{U_{rot}^{gt}})
    \end{equation}
    \begin{equation}
      \bm{L_{den}} = \bm{Lf} ( \bm{D}, \bm{U^{gt}})
    \end{equation}
    \begin{equation}
      \bm{L_{frot}} = \bm{Lf} ( \bm{\hat{U}_{rot}}, \bm{U_{rot}})
    \end{equation}
    
    The final loss \( \bm{L} \) is obtained by summing up the four aforementioned losses. We consider all four losses equally important and aim to control both prediction accuracy and rotation invariance simultaneously. Therefore, the weights $\alpha$, $\beta$, $\gamma$, $\delta$ for these four losses are all set to 1.
    \begin{align}
      \bm{L} = \bm{\alpha L_{ori}} + \bm{\beta L_{rot}} + \bm{\gamma L_{den}} + \bm{\delta L_{frot}}
    \end{align}

    \begin{table*}[t]
      \caption{Comparison with the State-of-the-art Methods on DENDI: Evaluation of F1-score}
      \label{tab2}
      \centering
      \small
      \setlength{\tabcolsep}{6pt} 
      \begin{tabular}{@{}c c cc ccc@{}} 
        \toprule
        \multirow{2}{*}{Method} & \multirow{2}{*}{F1-score (\%)} & \multicolumn{2}{c}{Metrics} & \multicolumn{3}{c}{Scores} \\
        \cmidrule(lr){3-4} \cmidrule(lr){5-7} 
        & & Rec. $\uparrow$ & Prec. $\uparrow$ & OE $\downarrow$ & PCC $\uparrow$ & Kappa $\uparrow$ \\
        \midrule
        SymResNet\cite{DBLP:conf/iccv/FunkL17} 
          & 30.70 & 0.4753 & 0.2431 & 0.1749 & 0.8251 & 0.2331 \\
        PMCNet\cite{DBLP:conf/iccv/SeoSC21} 
          & 52.00 & 0.6315 & 0.4430 & 0.1014 & 0.8986 & 0.4659 \\
        EquiSym\cite{DBLP:conf/cvpr/SeoKKC22}
          & 64.46 & 0.6608 & 0.6291 & 0.0636 & 0.9364 & 0.6097 \\
        ARDNet (Ours) & \bfseries{65.52} & \bfseries{0.6651} & \bfseries{0.6456} & \bfseries{0.0611} & \bfseries{0.9389} & \bfseries{0.6217} \\
        \bottomrule
      \end{tabular}
      
      \vspace{4pt}
      \footnotesize
      \begin{tabular}{@{}p{\textwidth}@{}}
        Note: Rec. = Recall, Prec. = Precision, OE = Omission Error, 
        PCC = Percentage of Correct Classification. 
        Bold indicates best results.
      \end{tabular}
    \end{table*}

    \begin{table*}[t]
      \caption{Comparison with the State-of-the-art Methods: Evaluation of Reflection Axis Detection}
      \label{tab3}
      \centering
      \setlength{\tabcolsep}{6pt} 
      \begin{tabular}{@{}c cccccc@{}} 
        \toprule
        \multirow{3}{*}{Dataset} & 
        \multicolumn{6}{c}{Accuracy Metrics} \\
        \cmidrule(lr){2-7}
        & \multicolumn{2}{c}{\makecell{Angular \\ Accuracy (\%) $\uparrow$}}
        & \multicolumn{2}{c}{\makecell{Center-Offset \\ Accuracy (\%) $\uparrow$}}
        & \multicolumn{2}{c}{\makecell{Overall \\ Accuracy (\%) $\uparrow$}}\\
        \cmidrule(lr){2-3} \cmidrule(lr){4-5} \cmidrule(lr){6-7}
        & EquiSym\cite{DBLP:conf/cvpr/SeoKKC22} & ARD(Ours) 
        & EquiSym & ARD 
        & EquiSym & ARD\\
        \midrule
        SDRW\cite{DBLP:conf/cvpr/LiuSZWPLRL13} 
          & 58.57 & \bfseries{64.29} & \bfseries{61.43} & 60.00 & 45.71 & \bfseries{54.29} \\
        LDRS\cite{DBLP:conf/iccv/SeoSC21}  
          & 22.50 & \bfseries{25.00} & 22.92 & \bfseries{27.08} & 12.5 & \bfseries{14.58} \\
        NYU\cite{DBLP:journals/prl/CicconetBLWG17}  
          & 67.78 & \bfseries{69.04} & 76.15 & \bfseries{78.66} & 60.25 & \bfseries{63.60} \\
        DENDI\cite{DBLP:conf/cvpr/SeoKKC22} 
          & 39.60 & \bfseries{55.45} & 42.57 & \bfseries{53.47} & 26.73 & \bfseries{40.59} \\
        \bottomrule
      \end{tabular}
      \vspace{4pt}
      \footnotesize
      \begin{tabular}{@{}p{\textwidth}@{}}
        Note: SDRW, LDRS, NYU, and DENDI contain 70, 240, 239, and 101 images respectively.
        Bold indicates best results.
      \end{tabular}
    \end{table*}

    \begin{table*}
      \caption{Comparison with the State-of-the-art Methods at Different Rotation Angles}
      \label{tab:freq}
      \centering
      \begin{tabular}{ccccccccccccc}
        \toprule
        \multirow{2}{*}{Angles}
        & \multicolumn{2}{c}{$ F1-score (\%) \uparrow$}
        & \multicolumn{2}{c}{$ Recall \uparrow $}
        & \multicolumn{2}{c}{$ Precision \uparrow $}
        
        & \multicolumn{2}{c}{$ OE \downarrow $}
        & \multicolumn{2}{c}{$ PCC \uparrow$}
        \\
        \cmidrule(lr){2-3} \cmidrule(lr){4-5} \cmidrule(lr){6-7}  \cmidrule(lr){8-9} \cmidrule(lr){10-11}  
        & Equi\cite{DBLP:conf/cvpr/SeoKKC22} & ARD(Ours) & Equi & ARD & Equi & ARD & Equi & ARD & Equi & ARD \\
        \midrule
        15° & 63.28  & \bfseries{65.65} & 63.18 &\bfseries{67.00} & 63.39  & \bfseries{64.36}
        & 0.0632 &  \bfseries{0.0604} & 0.9368 & \bfseries{0.9396} 
        \\
        30° & 60.34 & \bfseries{64.86} & 58.06 & \bfseries{64.91} & 62.81  & \bfseries{64.80}  
        & 0.0649 & \bfseries{0.0598} & 0.9351 & \bfseries{0.9402}
        \\
        45° & 57.79  & \bfseries{63.15} & 55.67 & \bfseries{63.37} & 60.09  & \bfseries{62.93}
        & 0.0676 & \bfseries{0.0615} & 0.9324 & \bfseries{0.9385} 
        \\
        60° & 58.77  & \bfseries{62.28} & 56.69 & \bfseries{61.84} & 56.69  & \bfseries{62.73}
        & 0.0649 & \bfseries{0.0611} & 0.9351 & \bfseries{0.9389}
        \\
        \bottomrule
      \end{tabular}
      \vspace{4pt}
      \footnotesize
      \label{tab4}
    \end{table*}

    \begin{table}
      \caption{Comparison with the State-of-the-art Methods on GMSYM}
      \label{tab:freq}
      \centering
      \begin{tabular}{ccccc}
        \toprule
        \multirow{2}{*}{scenarios}
        & \multicolumn{2}{c}{$ F1-score (\%) \uparrow$}
        & \multicolumn{2}{c}{$ Overall Accuracy (\%) \uparrow$}
        \\
        \cmidrule(lr){2-3} \cmidrule(lr){4-5}
        & Equi\cite{DBLP:conf/cvpr/SeoKKC22} & ARD(Ours) & Equi & ARD \\
        \midrule
        animal & 63.28  & \bfseries{65.65} & 62.18 & \bfseries{66.95} \\
        man & 60.34 & \bfseries{64.86} & 58.06 & \bfseries{64.91} \\
        building & 57.79  & \bfseries{63.15} & 55.67 & \bfseries{63.37} \\
        street sign & 58.77  & \bfseries{62.28} & 56.69 & \bfseries{61.84} \\
        flower & 62.68 & \bfseries{63.92} & 61.76 & \bfseries{62.51} \\
        cat & 61.27 & \bfseries{63.15} & 61.21 & \bfseries{63.56} \\
        fruit & 60.09  & \bfseries{62.73} & 56.69 & \bfseries{62.81} \\
        all & 62.38 & \bfseries{65.65} & 60.87 & \bfseries{64.24} \\
        \bottomrule
      \end{tabular}
      \vspace{4pt}
      \footnotesize
      \label{tab5}
    \end{table}

    \section{Experiments}

    \subsection{Experimental settings}
     \subsubsection{Datasets} We  evaluated our method on famous public symmetry detection dataset DENDI\cite{DBLP:conf/cvpr/SeoKKC22}, NYU\cite{DBLP:journals/prl/CicconetBLWG17}, LDRS\cite{DBLP:conf/iccv/SeoSC21} and SDRW\cite{DBLP:conf/cvpr/LiuSZWPLRL13}. 
     For reflection symmetries of continuous symmetry groups, DENDI represents an infinite number of line axes using elliptical masks.
    
     \subsubsection{New Symmetry Dataset GMSYM}
    Recently proposed DENDI dataset has a sufficient scale to support the training of deep learning models and annotates multiple reflection axes for individual images. However, existing reflection symmetry test datasets do not classify images and fail to consider complex images with interference. To address the aforementioned issues, we integrate diverse images based on DENDI, classify them under different scenarios, and apply various interferences to construct a new test dataset GMSYM.
    
     \subsubsection{Evaluation} To evaluate ARDNet, we use the F1-score as the primary evaluation metric. We compare the output score maps with ground-truth using algorithms such as morphological thinning and non-pixel matching, both at the pixel level. For the evaluation of reflection axis detection, we measure the angle $\theta$ between the detected symmetry axis($R$) and the ground-truth axis ($R_{GT}$). We also measure the distance $d$ from the center $c$ to the groundtruth line segment.

    \subsubsection{Implementation Details} 
    We employ a layer-wise upsampling strategy across each level of the backbone network, resulting in an equivariant UDecoder architecture. For training with rotated images, rotation angles are uniformly sampled from the range of 0° to 45°. All equivariant convolutions implemented in our framework are constructed based on the dihedral group D8. Both our focal loss \cite{DBLP:journals/pami/LinGGHD20} and rotation focal loss utilize a weighting factor $\alpha$ = 0.95 and a focusing parameter $\gamma$ = 2. We train the model for 100 epochs using the Adam optimizer with a learning rate of 0.001. The batch size is set to 16. Our proposed method is implemented with the PyTorch framework on four NVIDIA Tesla A100.

    \begin{figure*}[h]
    \centering
    \includegraphics[width=1.0\linewidth]{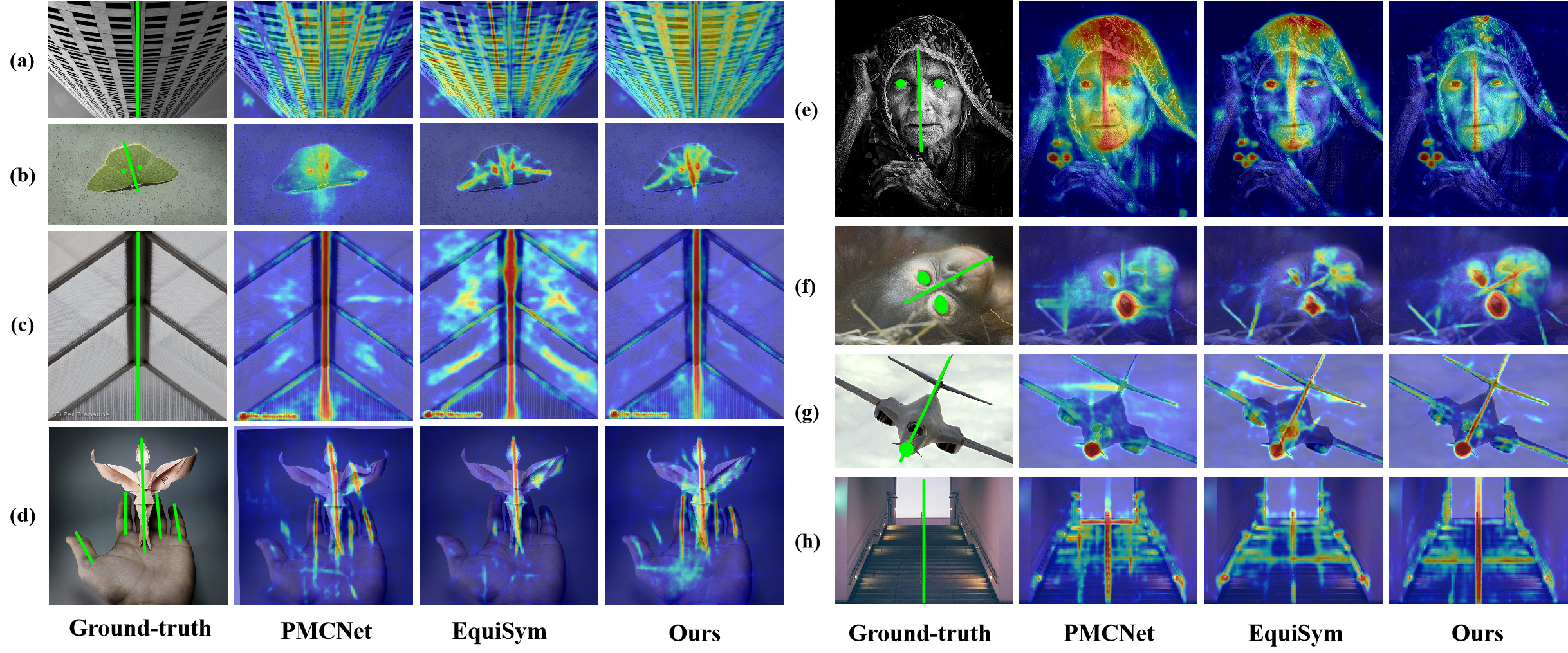}
    \caption{Qualitative results of the reflection symmetry detection on DENDI.}
    \label{fig4}
    \end{figure*}

    \begin{figure}[h]
    \centering
    \includegraphics[width=1.0\linewidth]{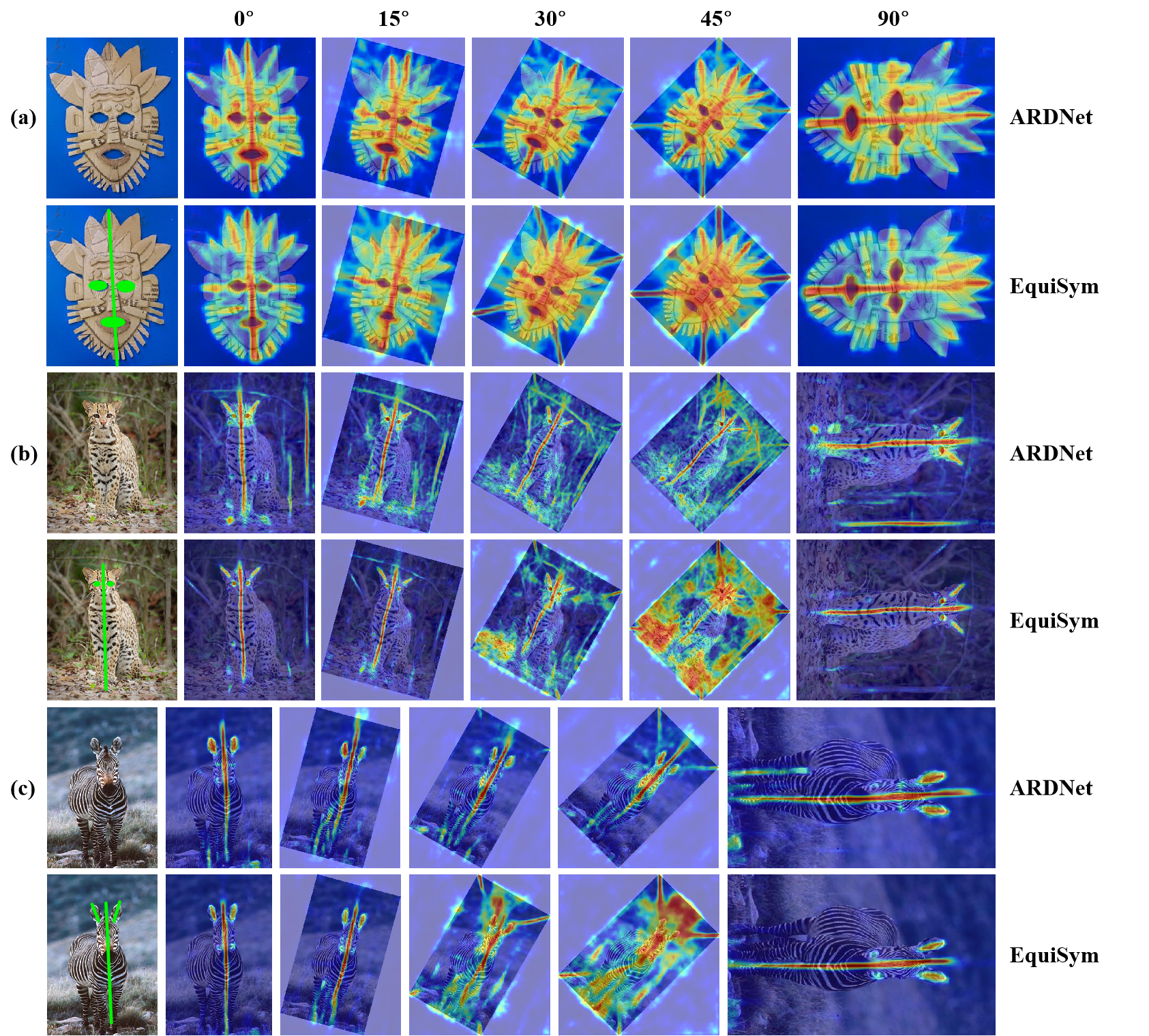}
    \caption{Rotation and perturbed images results of the reflection symmetry detection on DENDI.}
    \label{fig5}
    \end{figure}

     \subsection{Comparison with State-of-the-art Methods}
     We compare Our detection network with the state-of-the-art methods on DENDI in Tab. \ref{tab2} and Tab. \ref{tab3}. For reflection symmetry, our proposed ARDNet has reached state-of-the-art levels, demonstrating the effectiveness of asymmetric region denoising and rotation equivariant feature similarity matching. 
    
     Our method achieved an F1 score of 65.52\% in Tab. \ref{tab2}, significantly outperforming the state-of-the-art EquiSym at 64.49\%. We evaluate the angular accuracy, center-offset accuracy, and overall precision on four datasets, and our results in Tab. \ref{tab3} demonstrate superior performance across all three metrics compared to EquiSym.
     
     We also compare ARDNet with EquiSym on the newly proposed GMSYM dataset in Tab. \ref{tab5}. Experimental results demonstrate that ARDNet significantly outperforms EquiSym across diverse scenarios.
     
     
     To verify the effectiveness of ARDNet in detecting symmetric patterns in rotated images, we rotated the images by 15°, 30°, 45°, and 60° respectively on the DENDI dataset and compared the performance of ARDNet with that of EquiSym. The results in Tab. \ref{tab4} demonstrate that, compared to EquiSym, ARDNet exhibits excellent performance on rotated images at all angles. This supports our assertion that the rotation equivariant feature similarity matching is indispensable for rotation detection equivariance. Moreover, compared to EquiSym, our detection results focus more on symmetric regions with higher accuracy, indicating that asymmetric region denoising is an essential component.

    \subsection{Qualitative Results}
    
    The qualitative results of ARDNet, EquiSym and PMCNet on DENDI test are shown in Fig. \ref{fig4}. Both PMCNet and EquiSym fail to detect the actual symmetry axes, whereas ARDNet successfully identifies good symmetry.
    
    The qualitative results of rotated images are shown in Fig. \ref{fig5}. We compare ARDNet with EquiSym on DENDI datasets. We rotated the images by 15°, 30°, 45°, and 90° respectively to verify the rotation equivariance of ARDNet. ARDNet can accurately detect symmetric patterns in rotated images, demonstrating rotation robustness.
    
    To validate the robustness of the ARDNet, various interferences such as deformation, noise, and mosaic were introduced into the new dataset, as illustrated in Fig. \ref{fig5}. The ARDNet demonstrated strong performance even with the addition of these diverse disturbances.

     \subsection{Ablation Studies}
    We conducted extensive visual and statistical ablation experiments to validate the contribution of each component, as shown in Tab. \ref{tab1}. 
    Our study found that combining various losses and Rotation Equivariant Feature Similarity Matching with Asymmetric Region Denoising achieved the highest F1-score. During joint training, these two sub-tasks worked simultaneously and improved the accuracy of symmetry detection. This indicates that REFSM module not only enhances feature extraction capabilities but also ARD module accurately locates symmetric axis regions. The combined effect of these two modules increased our F1 score to 65.52\%.
    
    \begin{table}
      \caption{Ablation on the Symmetry Detection Network}
      \label{tab:freq}
      \centering
      \begin{tabular}{ccccccc}
        \toprule
            $\bm{L_{ori}}$ & $\bm{L_{rot}}$ & $\bm{L_{den}}$  & $\bm{L_{frot}}$ & REFSM & ARD &F1-score\\
        \midrule
         \checkmark &  &  &  & &  & 62.04\\
         \checkmark &\checkmark  &  &\checkmark  &\checkmark  &  & 63.55\\
         \checkmark & &\checkmark  &  & & \checkmark & 64.13\\
         \checkmark &\checkmark  &\checkmark  &\checkmark  &\checkmark & \checkmark  &  \bfseries{65.52}\\
      \bottomrule
    \end{tabular}
    \label{tab1}
    \end{table}

    Loss Weight Sensitivity Analysis. While our primary experiments employ equal weighting ($\alpha = \beta = \gamma = \delta = 1$) for all loss terms, we conduct a thorough sensitivity analysis to investigate the impact of varying these hyperparameters. We perform a grid search over $\alpha, \beta, \gamma, \delta \in \{0.5, 1.0, 2.0\}$ in Tab. \ref{tab6}, which presents a representative subset of configurations, including single-parameter variations and promising combinations.
    
    The results demonstrate that our framework is relatively robust to moderate weight variations, with F1-scores ranging from 64.12\% to 65.52\%. Increasing the rotation-related losses $\delta$ consistently yields performance gains, reaching a peak F1-score of 65.54\% at $\alpha = \beta = \gamma = \delta = 1 $. This validates the importance of rotation equivariance in our approach. Conversely, underweighting the denoising loss causes the most significant degradation to 64.12\%, confirming that asymmetric region denoising is critical for optimal performance. These findings suggest that while equal weighting provides strong results, emphasizing rotation consistency can yield marginal improvements.

    \begin{table}
      \caption{Sensitivity Analysis of Loss Weights on DENDI Dataset}
      \label{tab:sensitivity}
      \centering
      \begin{tabular}{ccccc}
        \toprule
        $\bm{\alpha\ (L_{ori})}$ & $\bm{\beta\ (L_{rot})}$ & $\bm{\gamma\ (L_{den})}$ & $\bm{\delta\ (L_{frot})}$ & F1-score (\%) \\
        \midrule
        0.5 & 1.0 & 1.0 & 1.0 & 64.87 \\
        2.0 & 1.0 & 1.0 & 1.0 & 65.03 \\
        1.0 & 0.5 & 1.0 & 1.0 & 64.45 \\
        1.0 & 2.0 & 1.0 & 1.0 & 65.12 \\
        1.0 & 1.0 & 0.5 & 1.0 & 64.12 \\
        1.0 & 1.0 & 2.0 & 1.0 & 65.48 \\
        1.0 & 1.0 & 1.0 & 0.5 & 64.93 \\
        \bfseries{1.0} & \bfseries{1.0} & \bfseries{1.0} & \bfseries{1.0} & \bfseries{65.52} \\
        \bottomrule
    \end{tabular}
    \label{tab6}
    \end{table}

    \section{CONCLUSION}
    Our paper addresses the challenges in reflection symmetry detection, particularly in interference from asymmetric regions and rotation non-equivariance. We propose utilizing the ARD module to denoise asymmetric regions and introduce Rotation-equivariant Feature Similarity Matching (REFSM). By maximizing the similarity between the score maps of rotated images and the ground truth, our approach effectively learns the symmetry detection patterns of rotated images, significantly enhancing the robustness of rotation prediction. Future work will further improve the accuracy of symmetry detection by exploring complex scenarios such as rotational variations and viewpoint changes.

\end{document}